%% file: acl2021.tex
\documentclass[11pt,a4paper]{article}
\usepackage[hyperref]{acl2021}
\usepackage{times}
\usepackage{latexsym}

\usepackage{microtype}

\aclfinalcopy
\def\aclpaperid{4256} 

\date{}

\input{math_commands.tex}

\usepackage{hyperref}
\usepackage{url}
\usepackage{graphicx}
\usepackage{subcaption}
\usepackage{soul}
\usepackage{xcolor}
\usepackage{caption}
\usepackage{subcaption}

\newcommand{\model}{CODS}

\title{
Controllable Abstractive Dialogue Summarization with \\
Sketch Supervision
}

\author{Chien-Sheng Wu\Thanks{\hspace{.2em}Equal contribution. Work mainly done when Linqing Liu was an intern at Salesforce Research. }\hspace{.3em}$^1$, 
Linqing Liu\footnotemark[1]\hspace{.1em} $^2$, Wenhao Liu$^1$,  Pontus Stenetorp$^2$, Caiming Xiong$^1$ \vspace{0.1cm}\\
$^1$ Salesforce Research $^2$ University College London \\
{\small \tt  \{linqing.liu,\hspace{.15em}p.stenetorp\}@cs.ucl.ac.uk, \{wu.jason,\hspace{.15em}wenhao.liu,\hspace{.15em}cxiong\}@salesforce.com}
}

\begin{document}
\maketitle
\begin{abstract} 
In this paper, we aim to improve abstractive dialogue summarization quality and, at the same time, enable granularity control.
Our model has two primary components and stages:
1) a two-stage generation strategy that generates a preliminary \textit{summary sketch} serving as the basis for the final summary.
This summary sketch provides a weakly supervised signal in the form of pseudo-labeled interrogative pronoun categories and key phrases extracted using a constituency parser.
2) A simple strategy to control the granularity of the final summary, in that our model can automatically determine or control the number of generated summary sentences for a given dialogue by predicting and highlighting different text spans from the source text.
%
Our model achieves state-of-the-art performance on the largest dialogue summarization corpus SAMSum, with as high as 50.79 in ROUGE-L score.
In addition, we conduct a case study and show competitive human evaluation results and controllability to human-annotated summaries.
%


\end{abstract}

\section{Introduction} 
Text summarization aims to produce an abridged version of the input text by distilling its most critical information.
In particular, abstractive -- as opposed to extractive -- summarization requires generative models with a high level of semantic understanding, as the output words do not necessarily appear in the source text.
While it is more challenging, it gives more flexibility to a summary compared to extractive summarization models~\citep{zhang2018neural}.
Significant research efforts have been focused on summarization of single-speaker documents such as text documents~\citep{liao2018abstract}, News~\citep{hermann2015teaching,nallapati2016abstractive,see2017get} or scientific publications~\citep{qazvinian2008scientific,nikolov2018data}.
However, dialogue summarization has not received much attention despite the prevalence of dialogues (text messages, email, social media, etc.) and the vast application potential of dialogue summarization systems.

Since dialogue language is inherently different from written text, it poses a unique set of challenges~\cite{zechner2001automatic}:
%
%
%
1) \textit{Distributed information across multiple speakers.} The most important information is usually scattered across several conversation turns from different speakers, while in articles it mostly presents in titles or the first few sentences.
2) \textit{Boundary detection.} In each turn pauses do not always match linguistic sensible segments; it is difficult to identify various critical information across turns due to surrounding non-content noise and disfluency.
3) \textit{Modeling interactions between speakers.} The speaker interaction plays an important role as it would imply the current dialog state and the status of the next speaker.
If we directly apply neural abstract summarization models which mostly encode the whole input only as a source sequence, the flow of the dialogue would be overlooked \cite{pan2018dial2desc}.
Previous methods \cite{goo2018abstractive, liu2019automatic} rely on explicit annotations to capture the logic of the dialogue, however, such annotations are not always available in datasets and additional labeling is cumbersome.

\begin{figure*}[t]
\begin{center}
\includegraphics[width=0.95\linewidth]{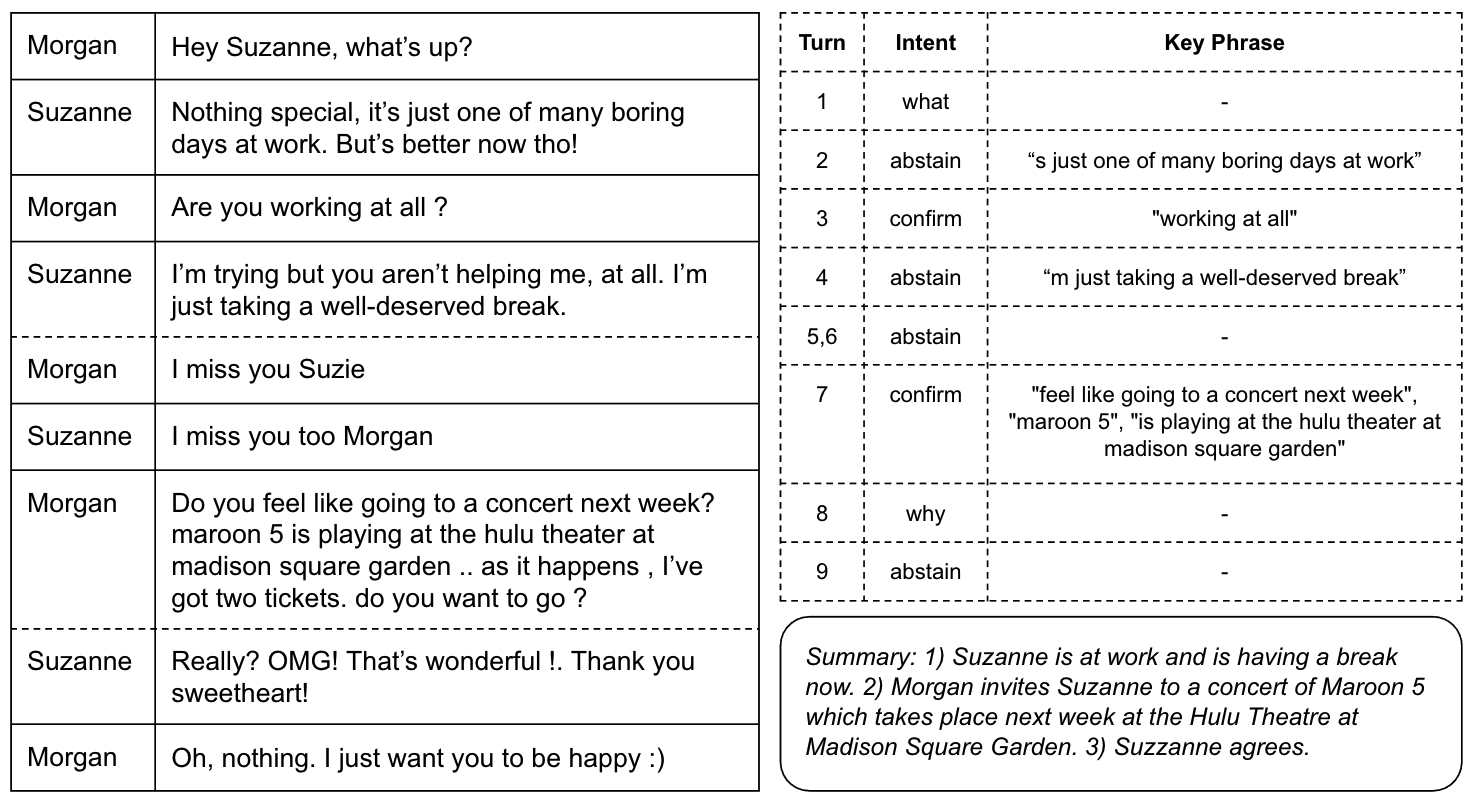}
\end{center}
\caption{An input and output example. Given the dialogue, we first construct a summary sketch with intent and key phrase information for each turn, and then split the dialogue into several segments (marked with dashed lines on the left hand side) for model controllability and interpretability.
}
\label{FIG:block_diagram}
\end{figure*}

To solve these challenges, we propose {\model}, a COntrollable abstractive Dialogue Summarization model equipped with sketch generation. 
%
We first automatically create a summary sketch that contains user intent information and essential key phrases that may appear in summary. It identifies the interaction between speakers and salient information in each turn. This summary sketch is prefixed to the human-annotated summary while fine-tuning a generator, which provides weak supervision as the final summary is conditioned on the generated summary sketch.
In addition, we propose a length-controllable generation method specifically for dialogue summarization. Desired lengths of summaries strongly depend on the amount of information contained in the source dialogue and granularity of information the user wants to understand \cite{kikuchi2016controlling}. We first segment the dialogue into different segments by matching each summary sentence linearly to its corresponding dialogue context. Then we train our model to generate only one sentence for each dialogue segment. This strategy makes use of the distributed information of the dialogue and make the generated summaries more trackable.


We base our model on BART-xsum~\citep{lewis2019bart}, which is first pre-trained with unsupervised denoising objectives, and further fine-tuned on the News summarization corpus XSUM~\citep{narayan2018don}.
We evaluate our approach on SAMSum~\citep{gliwa2019samsum}, the largest dialogue summarization dataset. Experimental results show that {\model} achieves state-of-the-art dialogue summarization performance on several automatic metrics. The main contributions of this work\footnote{Our code is released at \url{https://github.com/salesforce/ConvSumm}} are:
1) We propose a two-stage strategy that uses artificial summary sketch as weak supervision,
2) we introduce a text-span based conditional generation approach to control the granularity of generated dialogue summaries without human-written summaries at different detail levels, and
3) we conduct comprehensive case study and human evaluation to show that {\model} can achieve consistent and informative summary, especially for controllable summary, where existing models either cannot do it or do it poorly.


\section{Methodology}
Our model is based on pre-trained generative language models (Section~\ref{plms}).
Given an input dialogue history, our model first generates a summary sketch that serves as additional weakly supervised signal for the final summary (Section~\ref{draft_construction}).
Then it predicts the text span cutoffs over the entire dialogue and generates summaries accordingly (Section~\ref{control}).
%
We define the conversational history input as $D = \{X_1, X_2, \dots, X_N\}$, where each $X_i$ has a sequence of words, $N$ is the total numbers of dialogue turns, and the input may contain more than two speakers.
We intend to generate $M$-sentence dialogue summary $Y =\{Y_1,\dots,Y_M\}$ that is suppose to be briefer than the overall dialogue history. 
%

%
%
\subsection{Generative Pre-trained Language Models}
\label{plms}
As a first, our model needs transform a conversational history input into a dialogue summary.
%
%
Recently, self-supervised pretrained language models have
been employed as encoders and decoders since they~\citep{radford2019language,yang2019xlnet,dong2019unified} have achieved remarkable success across many NLP tasks. 
%
For general text summarization, this has also been the case with models such as BART~\citep{lewis2019bart} and PEGASUS~\citep{zhang2019pegasus}.
However, there are no results reported for self-supervised pretrained language models applied to dialogue summarisation, and people have argued that there is an intrinsic difference of linguistic patterns between human conversations and written text~\citep{wolf2019transfertransfo, wu2020tod, wu-2020-probing}.
%
We would like to answer the question which generative language model is the best base model for dialogue summarization tasks.


\subsection{Sketch Construction}
\label{draft_construction}
Conversational data, unlike news or scientific publications, includes lots of non-factual sentences such as chit-chats and greetings.
Removing these least critical information in the dialogues could potentially help the model better focus on the main content.
Based on this hypothesis, we combine a syntax-driven sentence compression method \cite{xu2019neural} with neural content selection.

Another potentially useful attribute for the conversational data is each dialogue turn inherently encodes user intent.
However, unlike task-oriented dialogue systems, which have explicit annotated intents (e.g., book flight and check account), dialogue summarization data rarely have such labels.
%
Thus we use a few heuristics with Snorkel~\citep{ratner2019training} to programmatically label each turn with a predefined interrogative pronoun category.
%
The generated intents and the compressed dialogues together constitutes the summary sketch as weakly-supervised signals.

To the best of our knowledge, in general, there is no non-task-oriented established label set.
Thus we draw upon the FIVE Ws principle, which often mentioned in journalism and research investigation, in that a passage can only be considered as complete if it answers these questions starting with such interrogative words \cite{5ws}. 
We adapt this principle to the dialogue scenario and identify a set of interrogative pronouns to support diverse enough user intents of all utterances, serving as the dialogue's logic.
For example, in Figure~\ref{FIG:block_diagram}, Morgan asked Suzanne ``Do you feel like going to a concert next week?'' One can expect that Suzanne will confirm her willingness in the next utterance. 
We define such dialogue intent categories including why, what, where, confirm, and abstain. More information for each category is shown in the Appendix (\ref{appn:da}).

To compress and remove noisy sub-sentences in the dialog, we first use a trained constituency parser~\citep{kitaev2018constituency} to parse each utterance. Then we compare the parsed phrases with the ground-truth summary to find their longest common sub-sequence (lcs), we set a threshold to filter and remove non-meaningful words (e.g., stop words) in lcs. Note that there are circumstances where the whole utterance is noisy and removable. 
Overall, we construct a summary sketch by concatenating utterance index, user intent label, and compressed utterance within the entire dialogue history into a string, ending with a special token, ``TL;DR". Take Figure \ref{FIG:block_diagram} as an example, the summary sketch is ``1 what 2 abstain 's just one of ... square garden 8 why 9 abstain TL;DR''. We train our model first to generate this summary sketch and then generate the final summary in an autoregressive way. We use TL;DR token to distinguish sketch and final summary during inference time.

\subsection{Controllability}\label{control}  
Due to the success of controllable language modeling~\cite{keskar2019ctrl}, the ability to control text summarization in the News domain has gradually been attracting attention~\citep{fan2018controllable,liu2018controlling}
The high-level intuition for our solution is that if we can control a generative model only to generate one sentence as output for a partially-highlighted input, we can control the number of output sentences by choosing how to highlight the input.
We highlight each dialogue split using the special token $<hl>$.
For example, in Figure~\ref{FIG:block_diagram}, we generate the first summary sentence for the first segment from turn one to four, and the second and third from turn five to seven and turn eight to nine, respectively (separated by the dashed lines). 
This way, we can not only gain the summary controllability but also make the generation more interpretable. 

The next challenge is, during training, we have to find a mapping between each sentence in a reference summary to its corresponding dialogue split. 
In other words, how do we know where to insert the highlighting tokens?
We do so by training a dialogue-turn-level binary classifier (detailed below) that predicts whether each turn is a cutting point (i.e., dialogue segmentation).
Our observation is that sentences within a reference summary usually have a strong temporal dependency, that is, people summarize the dialogue almost linearly. 
We use a simple approach to find the cutting points: the highest similarity score between conversations and each summary sentence. The cutting point
\begin{equation}
\begin{array}{c}
    t_m = \arg\max_{t} \textnormal{SIM}(X_{c_m:t}, Y_m),
\end{array}
\end{equation}
where SIM could be any similarity functions (we use ROUGE-1), and $c_m$ is the accumulated turn index ($c_1=1$ and $c_m=t_{m-1}$) that indicates which part of a dialogue has been covered. 
Note that for a summary with $M$ sentences, we only need to decide $M-1$ cutting points.
With the pseudo labels ($t_m$) provided by this heuristic, we formulate the dialogue segmentation problem into a binary classification problem. Specifically, we train a classifier $C$, which takes dialogue history as input and predicts whether each dialogue turn is a cutting point. We prefix each dialogue turn with a separation token as input to the classifier.
\begin{equation}
\begin{array}{c}
    H = C([x_{sep}, X_1, x_{sep}, X_2, \dots]) \in  \mathbb{R}^{N \times d_{emb}}, \\
    \hat{P} = \text{Sigmoid}(W_1(H)) \in \mathbb{R}^{N}.
\end{array}
\end{equation}
The classifier output $H$ is the representations of those separation tokens, and each of them is a $d_{emb}$ dimension vector. $W_1 \in \mathbb{R}^{d_{emb} \times 1}$ is a trainable linear mapping. The $\hat{P}$ is the predicted segment probability that is trained with binary cross-entropy loss. We use a BERT-base model~\citep{devlin2018bert} as classifier and the $i$-th cutting point is triggered if $\hat{P}_i > 0.5$. This prediction means that our model can automatically determine how many sentences should be generated in the final summary. If no cutting point is triggered, we generate a one-sentence summary. If one cutting point is triggered, we will have a two-sentence summary, and so forth. 

Finally, we can control the number of output summary sentences by controlling the dialogue split. Specifically, we first decide the expected number of output sentences (e.g., $K$), and then we choose the top $K-1$ indexes with highest probabilities in segmentation probability $\hat{P}$. We use these $K-1$ indexes as cutting points. 
We can also generate one-sentence summary by clipping the whole dialogue with one pair of highlighting tokens at the beginning and the end of a dialogue (we call this setting as {\model-1}).

\begin{figure}
     \centering
     \begin{subfigure}[b]{0.48\textwidth}
         \centering
         \includegraphics[width=\textwidth]{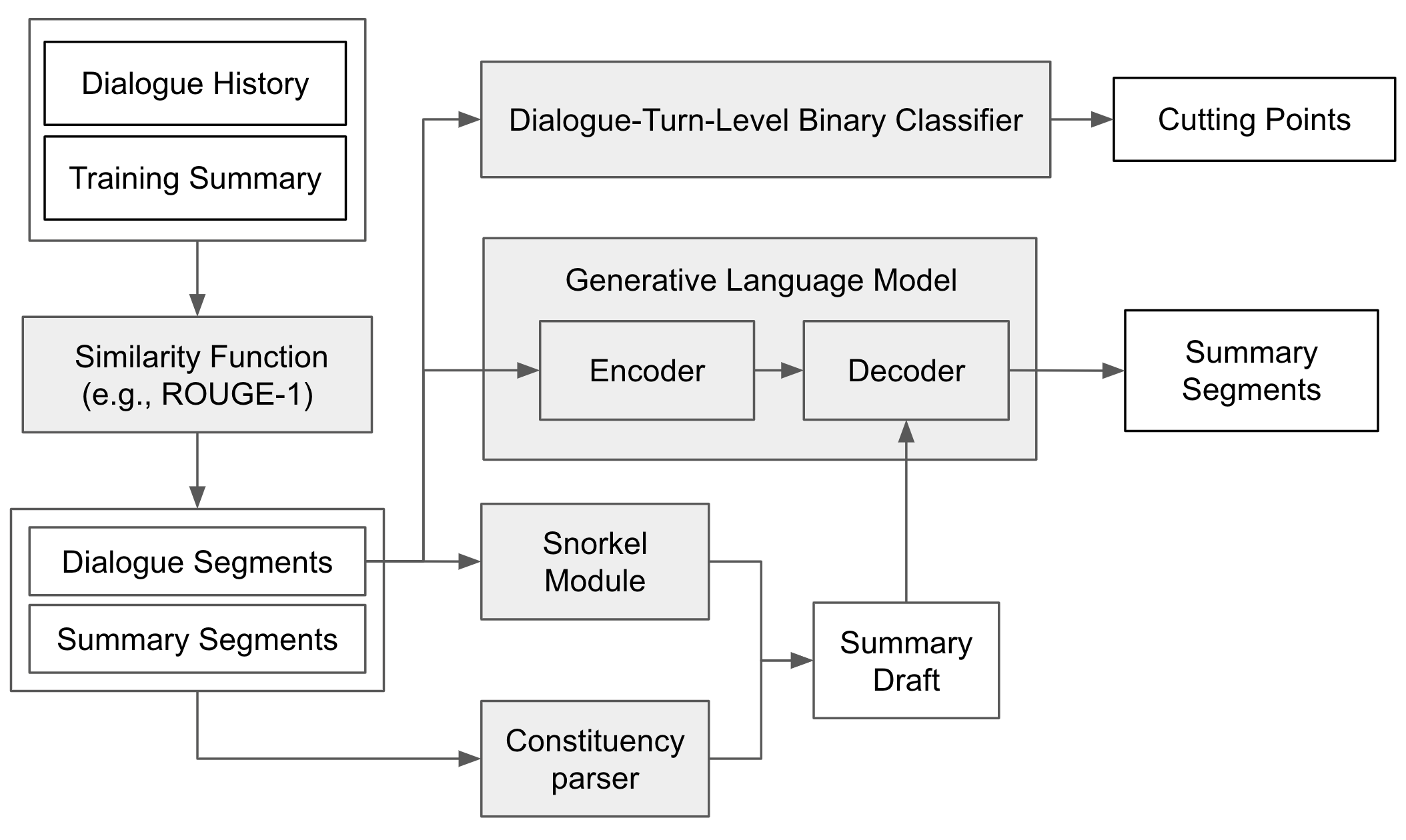}
         \caption{}
     \end{subfigure}
     \hfill
     \begin{subfigure}[b]{0.42\textwidth}
         \centering
         \includegraphics[width=\textwidth]{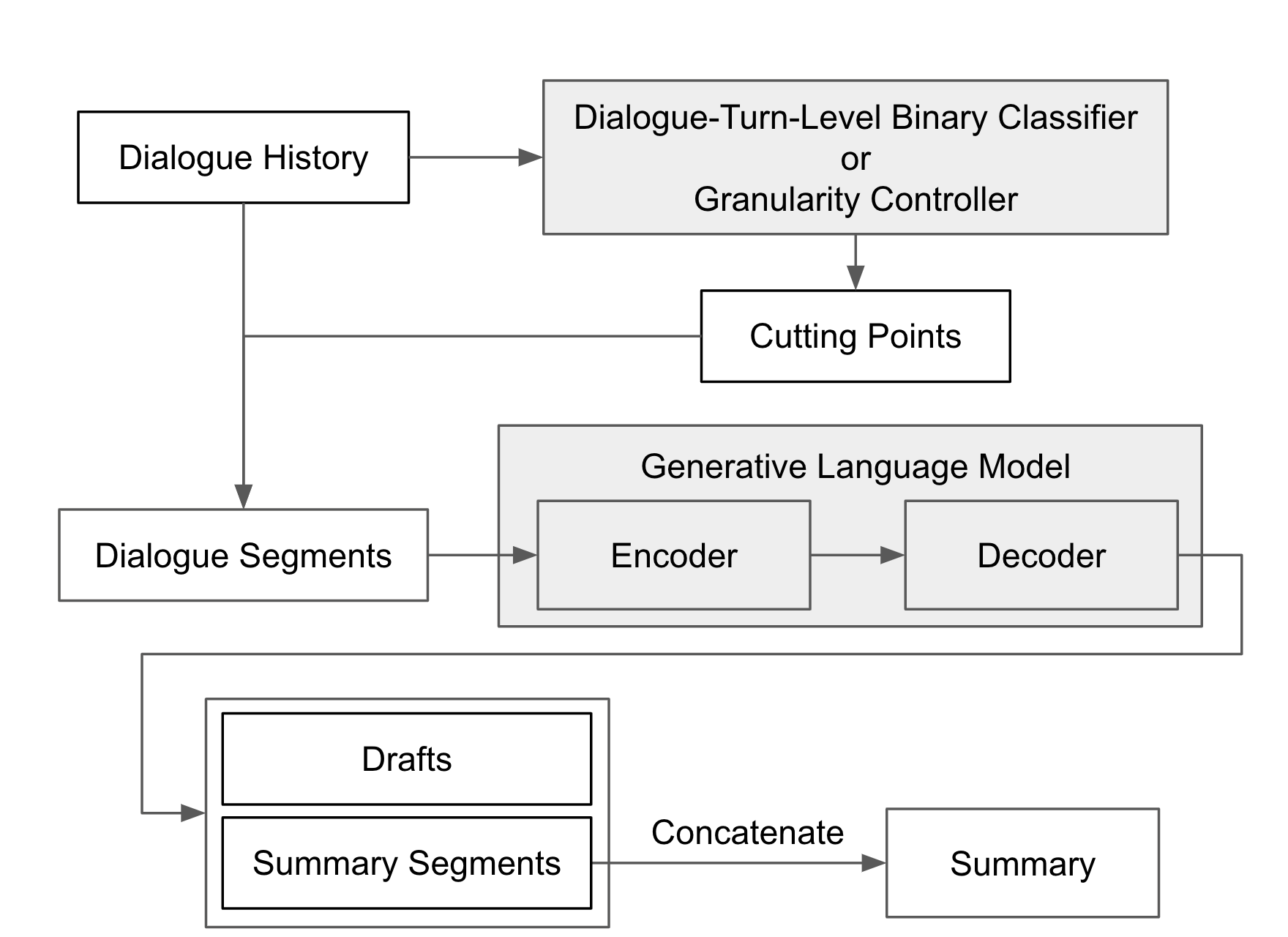}
         \caption{}
     \end{subfigure}
        \caption{(a) Training and (b) inference block diagrams of \model. Grey boxes are trainable functions.}
        \label{fig:two_blocks}
\end{figure}

\subsection{Overall Generation}
The overall training and inference block diagrams are shown in Figure~\ref{fig:two_blocks}. {\model} follows a standard encoder-decoder framework. During training, we use dialogue segmentation to add highlighting tokens for each summary sentence.
We take the highlighted dialogue history as input and train our model to generate its corresponding summary sketch and summary sentence. For example in Figure \ref{FIG:block_diagram}, the first summary sentence, we input the whole dialogue with added highlighting tokens both at the beginning of the first turn and at the end of the fourth turn, and generate output that contains the corresponding summary sketch ``1 what 2 abstain ... well-deserved break'' and the first summary sentence ``Suzanne is at work and is having a break now.''
The entire model is trained using cross-entropy loss for the generated tokens. During inference, we first use the trained binary classifier to predict cutting points. Then, we use the predicted segmentation to add highlighting tokens into a dialogue. Finally, after generating multiple summary sentences separately, we concatenate them to be the final summary.

\begin{table*}[t]
\begin{center}
\resizebox{0.8\linewidth}{!}{
\begin{tabular}{r|ccc}
\hline
\multicolumn{1}{l|}{}   & \textbf{ROUGE-1} & \textbf{ROUGE-2} & \textbf{ROUGE-L} \\ \hline
Longest-3*               & 32.46           & 10.27           & 29.92           \\
Pointer Generator~\citep{see2017get}*       & 37.27           & 14.42           & 34.36           \\
Fast Abs RL~\citep{chen2018fast}*             & 41.03           & 16.93           & 39.05           \\
Transformer~\citep{vaswani2017attention}*             & 42.37           & 18.44           & 39.27           \\
DynamicConv~\citep{wu2019pay}*             & 41.07           & 17.11           & 37.27           \\
DynamicConv + GPT-2 emb* & 45.41           & 20.65           & 41.45           \\
D-HGN~\citep{feng2020incorporating}         &      42.03 & 18.07 & 39.56       \\
TGDGA~\citep{zhao-etal-2020-improving}         &      43.11      &      19.15      &     40.49       \\
DialoGPT~\citep{zhang2019dialogpt}         & 39.77           &  16.58          &   38.42         \\
UniLM~\citep{dong2019unified}                    & 47.85           & 24.23           & 46.67           \\
PEGASUS~\citep{zhang2019pegasus}                  & 50.50           & 27.23           & 49.32           \\
BART-xsum~\citep{lewis2019bart}                & 51.74           & 26.46           & 48.72           \\ \hline
BART-xsum + Sketch (Ours)                & 51.79           & 26.85           & 49.15           \\
BART-xsum + Ctrl (Ours)                  & \textbf{52.84}           & 27.35           & 50.29           \\
{\model} (Ours)                 & 52.65  & \textbf{27.84}  & \textbf{50.79}  \\ \hline
\end{tabular}
}
\end{center}
\caption{Dialogue summarization ROUGE evaluation on the SAMSum test set. Results with * are obtained from \citealp{gliwa2019samsum}. {\model} achieves the highest ROUGE score. \textit{BART-xsum + Sketch} and \textit{BART-xsum + Ctrl} are ablated models individually removing controllability and sketch generation component from {\model}.} 
\label{TB:rouge}
\end{table*}

\begin{table*}[t]
\begin{center}
\resizebox{0.8\linewidth}{!}{
\begin{tabular}{r|cccccc}
\hline
\multicolumn{1}{l|}{} & ROUGE\_WE & BERTScore & MoverScore & BLEU & CIDEr & SMS \\ \hline
PEGASUS & 0.3562 & 0.5335 & 0.3233 & 17.33 & 1.741 & 0.1608 \\
BART-xsum & 0.3606 & 0.5387 & 0.3391 & 17.55 & 1.701 & 0.1401 \\
{\model} & \textbf{0.3759} & \textbf{0.5458} & \textbf{0.3539} & \textbf{19.58} & \textbf{1.981} & \textbf{0.1689} \\ \hline
\end{tabular}
}
\end{center}
\caption{Dialogue summarization evaluation on the SAMSum test set with additional recently introduced metrics that have been applied to both text generation and summarization.}
\label{TB:sumeval}
\end{table*}

\section{Experiments}

\subsection{Dataset}
We perform experiments on the recently released SAMSum dataset~\citep{gliwa2019samsum}~\footnote{The conversations in SAMSum may contain offensive words, please use the dataset carefully.}, which is the most comprehensive resource for abstractive dialogue summarization tasks. It contains 16K natural messenger-like dialogues created by linguists fluent in English with manually annotated summaries. This dataset is more challenging than the previous corpus \cite{kraaij2005ami} in the following aspects:
1) Unlike previous datasets consisting of only hundreds of dialogue-summary pairs, it has larger data size (16369 samples); 
2) 75\% of the conversations are between two interlocutors, the rest are between three or more people; 
3) the conversations cover diverse real-life topics, and the summaries are annotated with information about the speakers.
We preprocess the data by the following steps: 1) concatenate adjacent utterances of the same speaker into one utterance; 2) clean the dialogue text by removing hashtags, URLs and Emojis; 3) label each utterance with its corresponding interrogative pronoun category with a weak supervision approach \citep{ratner2019training}; 4) parse each utterance with a constituency parser and find the longest common sub-sequence between the phrases and summary to be the key phrases.

\subsection{Evaluation Metrics and Baselines}
We use the standard ROUGE metric~\citep{lin2004ROUGE} as automatic evaluation metrics, including ROUGE-1, ROUGE-2, and ROUGE-L. Following previous work~\citep{gliwa2019samsum}, we use py-ROUGE\footnote{\url{pypi.org/project/pyROUGE/}} library with stemming. We compare our model with baselines reported in \citealp{gliwa2019samsum}: Longest-3 is a commonly-used extractive summarization baseline which takes the top three longest sentences as summary. The pointer generator and Fast abs are RNN-based methods with copy-attention mechanism or policy gradient. The Transformer is a random-initialized self-attention architecture with multi-head attention. The DynamicConv is a lightweight convolutional model that can perform competitively to self-attention. All of these models are not pre-trained. 

Besides, we investigate four pre-trained generative language models to see which works the best for the dialogue summarization task.
DialoGPT is a GPT model pre-trained on open-domain Reddit data. 
UniLM is pre-trained using three types of language modeling tasks: unidirectional, bidirectional, and sequence-to-sequence prediction on English Wikipedia and BookCorpus. 
PEGASUS masks important sentences from input and is trained to generate the missing parts, similar to an extractive summary approach.
BART is trained by corrupting text with an arbitrary noising function and learning to reconstruct the original text.
We use default parameters listed in the respective open-source repositories to fine-tune on the dialogue summarization task. We show the training details in the Appendix.

\subsection{Results}
In Table \ref{TB:rouge} of ROUGE results, we find that the methods that are pre-trained or with pre-trained embeddings perform better than those that are not. For instance, DynamicConv achieves a 3 -- 4\% improvement by adding GPT-2 embeddings. This further confirms the impact of language model pre-training on downstream tasks. Among the pre-trained generative language models examined, PEGASUS and BART are the two top performance models with ROUGE-1 higher than 50. DialoGPT, the model pre-trained on conversational data, does not achieve satisfactory results,
implying that Reddit data has limited knowledge to be transferred to dialogue summarization tasks.
{\model} achieves the highest ROUGE score compared with other models, notably 50.79\% ROUGE-L.

To understand the individual contribution of each component in our model, we also conduct an ablation study by removing summary sketch generation (BART+Ctrl) or controllability (BART+Sketch). In both cases we observe a performance drop, except a slight improvement on ROUGE-1 for BART+Ctrl. This suggests that the sketching step helps generate a more fluent summary even with lower unigram matching. Furthermore, recognizing the limitation of ROUGE scores in their ability to fully capture the resemblance between the generated summary and the reference, in Table~\ref{TB:sumeval}, we follow \cite{fabbri2020summeval} to compare model performances with additional metrics, including ROUGE-Word Embedding~\citep{ng-abrecht-2015-better}, BERTScore~\citep{zhang2019bertscore}, MoverScore~\citep{zhao2019moverscore}, Sentence Mover’s Similarity (SMS)~\citep{clark2019sentence}, BLEU~\citep{papineni2002bleu}, and CIDEr~\citep{vedantam2015cider}. As shown in Table~\ref{TB:sumeval}, {\model} consistently outperforms PEGASUS and BART. More information about these evaluation metrics are shown in the Appendix.

\begin{table}[t]
\begin{center}
\resizebox{\linewidth}{!}{
\begin{tabular}{r|c|c|c}
\hline
\multicolumn{1}{l|}{} & Length Ratio & Consistent & Informative  \\ \hline 
Longest-1 & 0.27 & \textbf{0.70} & 0.23  \\ 
BART-xsum-1 & \textbf{0.16} & 0.50 & 0.16 \\ 
{\model}-1 & 0.19 & 0.50 & \textbf{0.49}  \\ \hline 
BART-xsum & 0.26 & 0.65 & 0.51  \\ 
{\model} & \textbf{0.24} & \textbf{0.69} & \textbf{0.53}  \\ 
Gold & 0.27 & 0.74 & 0.55  \\ \hline 
\end{tabular}
}
\end{center}
\caption{Human evaluation results on test set for both controllable summary and standard summary. }
\label{TB:humaneval}
\end{table}

\subsection{Analysis}

\begin{table*}[t]
\begin{center}
\resizebox{0.98\linewidth}{!}{
\begin{tabular}{|l|l|}
\hline
\multicolumn{2}{|l|}{Kelly: I still haven't received the rent money. Did you check with your bank?} \\
\multicolumn{2}{|l|}{John: Yes. I definitely sent it last week.} \\
\multicolumn{2}{|l|}{Kelly: But I still don't have it. Can you please check that you sent it to the right account.} \\
\multicolumn{2}{|l|}{John: Ok. Give me 5 min.} \\
\multicolumn{2}{|l|}{Kelly: OK} \\
\multicolumn{2}{|l|}{John: I checked and the money did go out of my account last week.} \\
\multicolumn{2}{|l|}{Kelly: What account number did you send it to?} \\
\multicolumn{2}{|l|}{John: 44-1278} \\
\multicolumn{2}{|l|}{Kelly: No wonder! My account number is 44-1279. You sent it to someone else's account.} \\
\multicolumn{2}{|l|}{John: ...! I'm really sorry!} \\
\multicolumn{2}{|l|}{Kelly: I still need the rent money though.} \\
\multicolumn{2}{|l|}{John: I'm really sorry I'll have to go to the bank tomorrow and ask if they can re-send it to the right account.} \\
\multicolumn{2}{|l|}{Kelly: Thanks !} \\ \hline
\multicolumn{1}{|r|}{Longest-1} & John said I'm really sorry I'll have to go to the bank tomorrow and ask if they can re-send it to the right account. \\ \hline
\multicolumn{1}{|r|}{BART-1} & Kelly still hasn't received the rent money from John. \\ \hline
\multicolumn{1}{|r|}{{\model}-1} & \begin{tabular}[c]{@{}l@{}}John sent the rent money to the wrong account and will have to ask the bank to re-send it to the correct \\ one tomorrow.\end{tabular} \\ \hline
\multicolumn{1}{|r|}{BART} & \begin{tabular}[c]{@{}l@{}}Kelly still hasn't received the rent money. John sent it to the wrong account number 44-1278. John will go \\ to the bank tomorrow and ask if they can re-send the money to the right account.\end{tabular} \\ \hline
\multicolumn{1}{|r|}{\model} & \begin{tabular}[c]{@{}l@{}}\textbf{Sketch}: 1 \#confirm haven't received the rent money check with your bank 2 none 3 \#confirm check that you sent it to \\ the right account 4 none 5 none 6 \#abstain the money did go out of my account last week 7 \#abstain did you send it to \\ 8 none 9 \#what sent it to someone else's account 10 none 11 \#abstain need the rent money though 12 \#abstain 'm really \\ sorry i'll have to go to the bank tomorrow and ask if they can re-send it to the right account 13 none\\  \textbf{Summary}: John sent the rent money to the wrong account last week. John will go to the bank tomorrow and ask if he \\ can re-send the money to the correct account.\end{tabular} \\ \hline

\multicolumn{1}{|r|}{Gold} & \begin{tabular}[c]{@{}l@{}}Kelly hasn't received the rent money, because John sent it to the wrong bank account. He will go to the \\ bank to tackle the issue.\end{tabular} \\ \hline
\end{tabular}
}
\end{center}
\caption{A test set example with generated summaries.}
\label{TB:example1}
\end{table*}

\begin{table*}[t]
\centering
\begin{center}
\resizebox{0.98\linewidth}{!}{
\begin{tabular}{|l|l|l|l} 
\cline{1-3} & \multicolumn{1}{c|}{Reference Summary}  & \multicolumn{1}{c|}{{\model} Summary}&  \\ 
\cline{1-3}
\multicolumn{1}{|c|}{\begin{tabular}[c]{@{}c@{}}Associate names \\with actions\end{tabular}}  & \begin{tabular}[c]{@{}l@{}}Lilly will be late. \\\textcolor{blue}{Gabriel} will order pasta with salmon and basil for \textcolor{blue}{her}.\end{tabular} & \begin{tabular}[c]{@{}l@{}}Lilly will be late for the meeting with Gabriel. \\\textcolor{blue}{Gabriel} will order something for \textcolor{blue}{Lilly}.\end{tabular} & \\ 
\cline{2-3}
\multicolumn{1}{|c|}{}& \begin{tabular}[c]{@{}l@{}}Ann doesn't know what she should give to \textcolor{blue}{her dad} as a birthday gift. \\He's turning 50. \\\textcolor{blue}{Fiona tries to help her} and suggests a paintball match.\end{tabular} & \begin{tabular}[c]{@{}l@{}}It's \textcolor{blue}{Ann's dad}'s 50th birthday. \\He's turning 50. \textcolor{blue}{Ann and Fiona} are planning a \\surprise birthday party for her dad.\end{tabular} & \multicolumn{1}{c}{}  \\ 
\cline{1-3}
\begin{tabular}[c]{@{}l@{}}Extract information after\\the discussion\end{tabular}  & Paul will buy \textcolor{blue}{red roses }following Cindy's advice.& Paul wants to buy \textcolor{blue}{ red roses.}&  \\ 
\cline{1-3}
\multicolumn{1}{|c|}{\begin{tabular}[c]{@{}c@{}}Decide important \\information\end{tabular}} & \begin{tabular}[c]{@{}l@{}}Rachel's aunt had an accident and she's in hospital now.\\She's only bruised. \\\textcolor{red}{The perpetrator of the accident is going to pay for the rehabilitation.}\end{tabular} & \begin{tabular}[c]{@{}l@{}}Rachel is at the hospital with her aunt, \\who had an accident. \\She's bruised but fine. \\She will give her a hug.\end{tabular} & \multicolumn{1}{c}{}  \\ 
\cline{2-3}
& \begin{tabular}[c]{@{}l@{}}\textcolor{red}{Hannah needs Betty's number} but Amanda doesn't have it. \\She needs to contact Larry.\end{tabular} & \begin{tabular}[c]{@{}l@{}}Amanda can't find Betty's number. \\Amanda suggests to text him.\end{tabular} 
&\\
\cline{1-3}
\end{tabular}
}
\end{center}
\caption{Case analyses by manually examining {\model} generated summaries.}
\label{Tab: case_analyses}
\end{table*}

\subsubsection{Human Evaluation by Crowdsourcing}

We leverage human judgement to evaluate the generated summaries via crowdsourcing, especially for granularity-controlled generation, since we do not have human-written reference summaries of various lengths (number of sentences).
We ask workers to rate the summaries in two aspects on a scale from -1 (worst) to 1 (best): factual consistency and informativeness. \emph{Factual consistency} acts as a precision measure, assessing whether the information provided in summary contains factual errors which are against the source dialogue;
\emph{Informativeness} is a recall-oriented measure, examining whether critical information in a dialogue is mentioned in summary.
We also show the length ratio between a summary and a dialogue, where a lower ratio means a higher compression rate. For the crowdsourcing evaluation, we randomly select 6\% dialogues from the test set, each of which is annotated by three workers. More details about human evaluation process are in the Appendix~\footnote{The prediction file on the test set is provided in the supplementary file.}.

To show the proposed controllable generation's strengthens and quality, we provide two additional baselines, Longest-1 and BART-1. The longest-1 method is an extractive baseline that outputs the longest dialogue turn as the final summary. The BART-1 is a strong abstractive baseline where we train a BART-based summarization model with the number of summary sentences in the training set as its start-of-sentence token during decoding. Similar to the approach from \citealp{liu2018controlling}, we can use different start-of-sentence tokens to control the BART output.

In general, it is preferable to have a factually consistent and informative summary that is succinct (low length ratio, high compression rate) at the same time. 
As shown in the first row of Table~\ref{TB:humaneval}, {\model-1} achieves the highest informative score among all generated one-sentence summaries, indicating the strength of the proposed controllable method in producing succinct yet informative dialogue summaries. 
The Longest-1 method has a higher consistent score because its summary is directly copied from the original dialogue, preventing any factual mistakes.
The second row of Table~\ref{TB:humaneval} shows that {\model}, when automatically determining the granularity of the summary, produces summaries that are more succinct (lower length ratio), more factually consistent, and more informative, compared to the BART model.

\subsubsection{Case Study}
{\model} outperforms the baseline models in both ROUGE scores and human evaluation metrics. We now further inspect its textual quality. In Table~\ref{TB:example1}, we show an example from the SAMSum test set with summaries generated by different models.
In this example, {\model} and {\model-1} can both produce a near-perfect summary even compared to the human-written reference summary.
On the other hand, the summary generated by BART includes overly detailed information (e.g., bank account).
We show some more examples in the Appendix and all the predictions (including {\model-1} and {\model-2}) in the supplementary file.

We also manually examine 100 summaries generated from {\model} against the reference summaries in the test set.
Specifically, we analyze each of the three following problematic cases, where summarization models frequently make mistakes, reported by \citealp{gliwa2019samsum}, and provide sample summaries in Table \ref{Tab: case_analyses}.
1) \textit{Associating names with actions}: {\model} performs well in dealing with speakers' names. It accurately associates ``her dad" with ``Ann's dad," also ``Fiona tries to help her" with ``Ann and Fiona." 2) \textit{Extract information about the arrangement after discussion}: Even speakers hesitate about the flower's color to be yellow, pink or red in the middle of the discussion, {\model} still correctly determines the right color after several turns. 3) \textit{Decide important information in dialogues}: {\model} fails to capture some of the important facts (marked as red) mentioned in reference summary.
We conjecture the reason could be that 1) some of the important facts are located in the same part of the highlighted turns, and 2) those information is missed by the key phrase extraction.
Simultaneously, we force the model to generate only the most important one under the constraint of controllability. The improvement of {\model} on the first two summarization difficulties can be partially attributed to the clear logic in the sketch when input to the model.

\section{Related Work}
\paragraph{Neural Text Summarization}
There are two main paradigms for text summarization: extractive and abstractive. 
Inspired by the success of applying seq2seq models on neural machine translation, \citealp{rush2015neural} and \citealp{nallapati2016abstractive} introduce the neural seq2seq model on abstractive text summarization, with an attention-based encoder and a neural language model decoder. 
To solve the problem of out-of-vocabulary words and to capture salient information in source documents, \citealp{see2017get} propose a pointer-generator network that copy words from source to target. 
Many subsequent works \citep{gehrmann2018bottom, paulus2018deep} 
further demonstrate its effectiveness with reinforcement learning. 
Recently, \citealp{liu2019text} apply BERT on text summarization and propose a general framework for both extractive and abstractive models. \citealp{zhang2019hibert} pre-train hierarchical document encoder for extractive summarization. 
\citealp{lewis2019bart} introduces BART, a denoising autoencoder for pretraining sequence-to-sequence models. BART significantly outperforms the best previous work in terms of ROUGE metrics. 

\paragraph{Dialogue Summarization}
Regarding to the datasets in dialogue summarization, initial abstractive dialogue summarization work~\citep{oya2014template, mehdad2014abstractive, banerjee2015abstractive} are conducted on the AMI meeting corpus~\citep{kraaij2005ami}, with only 141 summaries.
\citealp{goo2018abstractive} propose to use the topic descriptions (high-level goals of meetings) in AMI as reference summaries and use dialogue acts as training signals.
\citealp{pan2018dial2desc} build the Dial2Desc dataset by reversing a visual dialogue task, aligning image dialogues with the image caption as a summary.
\citealp{liu2019automatic} collect their dataset from the logs in the DiDi customer service center. It is restricted to task-oriented scenario, where one speaker is the user and the other is the customer agent, with limited topics and it is also connected to the goal of dialogue state tracking task~\citep{wu-etal-2019-transferable,wu-etal-2020-improving-limited}.
Recently, \citealp{gliwa2019samsum} introduce the SAMSum corpus, with 16k chat dialogues with manually annotated summaries. It is the first comprehensive abstractive dialogue summarization dataset spanning over various lengths and topics. \citealp{chen2020multi} propose a multi-view sequence-to-sequence model by extracting different views
of structures from conversations. Both their method and ours leverage rich conversation structure information. Evaluating on SAMSum, our model {\model} outperform theirs by 3 points in terms of ROUGE scores, indicating our utilized dialogue features are more effective.

\paragraph{Length-controllable Generation}
The most prevalent method for length control generation is using a special length embedding. \citealp{kikuchi2016controlling} first propose length control for abstractive summarization by using length embedding as an additional input for the LSTM. \citealp{fan2018controllable} train embeddings that correspond to each different output length and prepend that length marker at the beginning of the decoder. \citealp{liu2018controlling} incorporates the length embedding into initial state of a CNN-based decoder. \citealp{takase2019positional} extends the positional encoding in Transformer model by considering the remaining length explicitly at each decoding step. \citealp{saito2020length} propose to control the summary length with prototype extractor.
However, the retrieve-and-rewrite process is restricted by the extraction quality, leaving its performance limited by extractive solutions' capabilities.
The aforementioned works all focus on structured text summarization (e.g. news document). We are the first to propose generate length-controllable summary on dialogues by highlighting arbitrary numbers of dialogue spans.

\section{Conclusion}
The dialogue summarization task is challenging but with vast application potential. We propose {\model}, a state-of-the-art dialogue summarization model with granularity controllability. {\model} uses a weakly-labeled summary sketch for its two-stage generation, and text-span conditional generation for a controllable summary. Our model surpasses existing models on the largest dialogue summarization dataset. 
We show with human evaluation that our model can generate factually consistent and informative summaries.
We also point out several error cases to shed light on future research direction of controllable dialogue summarization.



\bibliographystyle{acl_natbib}
\bibliography{anthology,acl2021}

\clearpage
\newpage
\appendix
\section{Appendix}
\input{appendix}

\end{document}

%% file: math_commands.tex
\usepackage{amsmath,amsfonts,bm}

\def\eqref#1{equation~\ref{#1}}

\def\1{\bm{1}}

\DeclareMathAlphabet{\mathsfit}{\encodingdefault}{\sfdefault}{m}{sl}
\SetMathAlphabet{\mathsfit}{bold}{\encodingdefault}{\sfdefault}{bx}{n}



%% file: appendix.tex
\subsection{}
\label{appn:da}
We define dialogue intent categories as follows: \textit{WHY:} asks the reason of the state mentioned in the previous turn, e.g., ``why'' or ``why not''; \textit{WHAT:} requests more details about the aforementioned object, the sentence usually starts with ``what's'' or ``what about''; \textit{WHERE:} the location of the event; \textit{WHEN:} the time of the event, e.g. ,``when'' or ``what time''; \textit{CONFIRM:} expects the other speaker to establish the correctness of a certain case, the sentence usually starts with patterns like ``are you'', ``will you'' or ``has he''; \textit{ABSTAIN:} the utterance does not belong to any of the previous categories. It occurs when speakers continue to state or comment without seeking for more information from the others.

\subsection{}
\begin{itemize}
    \item DialoGPT: A GPT model pretrained on 147M conversation-like data extracted from Reddit comments. We use the model with 117M parameters. \url{github.com/microsoft/DialoGPT} 
    \item UniLM: A multi-layer Transformer network optimized for three language modeling objectives: unidirectional, bidirectional and sequence-to-sequence prediction. It is initialized with BERT$_{\text{LARGE}}$, then pre-trained using English Wikipedia and BookCorpus. Same as BERT$_{\text{LARGE}}$, it contains 340M parameters.
    \url{github.com/microsoft/unilm}
    \item PEGASUS: They pretrain a Transformer-based encoder-decoder models with a new self-supervised objective - gap-sentence generation - on the C4 corpus. We use the PEGASUS of 568M parameters. 
    \url{github.com/google-research/pegasus}
    \item BART: Transformer-based encoder-decoder model trained by corrupting text with an arbitrary noising function and learning a model to reconstruct the original text. We use BART$_{\text{LARGE}}$ model which contains 400M parameters.
    \url{huggingface.co/transformers/model_doc/bart.html} 
    \item \model: It's based on BART$_{\text{LARGE}}$ model which contains 400M parameters.
\end{itemize}

\begin{table}
\begin{center}
\resizebox{\linewidth}{!}{
\begin{tabular}{|l|l|}
\hline
\multicolumn{2}{|l|}{Keith: Meg, \textcolor{blue}{pls buy some milk and cereals}, I see now we've run out of them .} \\
\multicolumn{2}{|l|}{Megan: hm, sure, I can do that .} \\
\multicolumn{2}{|l|}{Megan: but did you \textcolor{blue}{check in the drawer next to the fridge} ?} \\
\multicolumn{2}{|l|}{Keith: nope, let me have a look .} \\
\multicolumn{2}{|l|}{Keith: ok, false alarm, we \textcolor{blue}{have cereal and milk} .} \\
\multicolumn{2}{|l|}{Deana: glad to hear it !} \\ \hline
Summary & \multicolumn{1}{l|}{Megan needn't buy milk and cereals. They're in the drawer next to the fridge.} \\ \hline
\end{tabular}}
\end{center}
\caption{Example key phrases in summary sketch.}
\label{TB: sketch_exp1}
\end{table}

\begin{table}
\begin{center}
\resizebox{\linewidth}{!}{
\begin{tabular}{|l|l|}
\hline
\multicolumn{2}{|l|}{Norbert: we \textcolor{blue}{need to hurry to catch the tour} .} \\
\multicolumn{2}{|l|}{Wendy: ok , \textcolor{blue}{am buying something} . be right out !} \\
\multicolumn{2}{|l|}{Norbert: ok . am not waiting long though . missed the last one because of you .} \\
\multicolumn{2}{|l|}{Wendy: just be patient for once .} \\
\multicolumn{2}{|l|}{Norbert: im always patient .} \\
\multicolumn{2}{|l|}{Wendy: at the register now .} \\ 
\multicolumn{2}{|l|}{Norbert: alright .} \\ \hline
Summary & \multicolumn{1}{l|}{Wendy is shopping, but she needs to hurry up to catch the tour.} \\ \hline
\end{tabular}}
\end{center}
\caption{Example key phrases in summary sketch.}
\label{TB: sketch_exp2}
\end{table}

\begin{table}
\begin{center}
\resizebox{\linewidth}{!}{
\begin{tabular}{|l|l|}
\hline
\multicolumn{2}{|l|}{Phil: \textcolor{blue}{is} brandon in ?} \\
\multicolumn{2}{|l|}{Clara: not yet .} \\
\multicolumn{2}{|l|}{Phil: has he \textcolor{blue}{called to say he'd be late} ?} \\
\multicolumn{2}{|l|}{Clara: no , he \textcolor{blue}{hasn't} .} \\
\multicolumn{2}{|l|}{Phil: it's not the first time , ist it ?} \\
\multicolumn{2}{|l|}{Clara: no , it isn't .} \\ 
\multicolumn{2}{|l|}{Phil: when he arrives , tell him to come to me .} \\
\multicolumn{2}{|l|}{Clara: no , it isn't .} \\
\multicolumn{2}{|l|}{Phil: please \textcolor{blue}{prepare a report on the absenteeism and lateness . i expect it by friday on my desk} .} \\ 
\multicolumn{2}{|l|}{Clara: it \textcolor{blue}{will be} ready .} \\ 
\hline
Summary & \multicolumn{1}{l|}{\begin{tabular}[c]{@{}l@{}}Brandon is late again. Clara will prepare a report on the absenteeism \\ and lateness for Phil by Friday.\end{tabular}} \\ \hline

\end{tabular}}
\end{center}
\caption{Example key phrases in summary sketch.}
\label{TB: sketch_exp3}
\end{table}

\subsection{Sketch Construction}
Previous methods~\citep{goo2018abstractive, pan2018dial2desc} heavily rely on explicit intent annotations in datasets. We label user intent automatically for each utterance with the Snorkel library in a weak supervision approach. For each interrogative pronoun category, we first manually identify its most frequent key words and patterns (can be found in our source code). Then we use the labeling functions in Snorkel to label all the utterances. 

For the utterance compression, we do LCS on the phrases generated from the constituency parser. In the example of \ref{FIG:block_diagram}, \textit{s just one of many boring days at work}  the parsed constituent overlapping with ‘at work’ in the summary, so we keep this phrase. However, in other examples, not all overlapped words are meaningful (e.g. stop words). We thus filter the LCS results and only keep important key phrases. Then we train our model to predict these key phrase spans in each turn. We show three examples of our generated key phrases in summary sketches on evaluation set (see Table \ref{TB: sketch_exp1}, \ref{TB: sketch_exp2}, \ref{TB: sketch_exp3})

\subsection{Training Details}
We use huggingface~\citep{wolf2019transformers} implementation to fine-tune a BART model. We use the large version fine-tuned on the XSUM~\citep{narayan2018don} dataset with 12 self-attention encoder and decoder layers. We truncate input dialogue to a maximal length 512 with training batch size 4. We train the model with Adam optimizer~\citep{kingma2014adam} with 0.1 proportion for linear learning rate warmup. We early stop on validation set ROUGE-1 score, and it is trained for around 40,000 steps on one NVIDIA V100 GPU. During inference, we do beam search decoding with beam size 4.

\subsection{Evaluation Metrics}
Information obtains from \cite{fabbri2020summeval}:
\begin{itemize}
    \item ROUGE measures the number of overlapping textual units between the generated summary and a set of reference summaries.
    
    \item ROUGE-WE extends ROUGE by taking cosine similarity of Word2Vec embeddings into account.
    
    \item BERTScore computes similarity scores by aligning generated and reference summaries on a token-level based on the output of the BERT-based model. Token alignments are computed greedily with the objective of maximizing the cosine similarity between contextualized token embeddings. We report the F1 score.
    
    \item MoverScore measures semantic distance between a summary and reference text by making use of the Word Mover’s Distance operating over n-gram embeddings pooled from BERT representations.
    
    \item Sentence Mover’s Similarity (SMS) extends Word Mover’s Distance to view documents as a bag of sentence embeddings as well as a variation which represents documents as both a bag of sentences and a bag of words.
    
    \item BLEU is a corpus-level precision-focused metric which calculates n-gram overlap between a candidate and reference utterance and includes a brevity penalty. It is the primary evaluation metric for machine translation.

    \item CIDEr computes {1-4}-gram co-occurrences between the candidate and reference texts, down-weighting common n-grams and calculating cosine similarity between the ngrams of the candidate and reference texts.
\end{itemize}

\subsection{Human Evaluation}
We use roughly 6\% of the test set data in SAMSum for human evaluation and we do some filtering based on the annotation of the “gold summary”. Specifically, we filter those annotations if a “gold summary” has been annotated as “-1” (the meaning of each score is shown below), implying that the annotators may not pay attention to the scoring. The final results reported in Table 3 is the mean from three different annotators.

The “gold summary” is actually not perfect and it might contain some noisy annotation, this is the reason why some workers may give 0 even if it is collected from humans. Below is the scoring instruction we sent to our workers:
\begin{itemize}
    \item Factual Consistency (Precision): The rating measures whether the information provided in a summary is correct. Score -1 if a summary contains a serious factual error. Score 0 if a summary has some minor factual errors. Score 1 if everything in a summary is factually correct.
    \item Informative (Recall): The rating measures whether all the important information in a dialogue is included in a summary. Score -1 if a summary misses serious key points. Score 0 if a summary misses a few key points. Score 1 if a summary covers all key points.
\end{itemize}




\begin{table*}[h]
\begin{center}
\resizebox{0.7\linewidth}{!}{
\begin{tabular}{|l|l}
\hline
\multicolumn{2}{|l|}{Paul: what color flowers should i get} \\
\multicolumn{2}{|l|}{Cindy: any just not yellow} \\
\multicolumn{2}{|l|}{Paul: ok , pink ?} \\
\multicolumn{2}{|l|}{Cindy: no maybe red} \\
\multicolumn{2}{|l|}{Paul: just tell me what color and what type ok ?} \\
\multicolumn{2}{|l|}{Cindy: ugh , red roses !} \\
\hline
\multicolumn{1}{|r|}{Gold} & \multicolumn{1}{l|}{Paul will buy red roses following Cindy's advice.} \\  \hline
\multicolumn{1}{|r|}{BART} & \multicolumn{1}{l|}{Paul wants to get red roses. Cindy doesn't want pink or yellow roses.} \\ \hline
\multicolumn{1}{|r|}{\model} & \multicolumn{1}{l|}{Paul wants to buy red roses.} \\ 
\hline
\end{tabular}
}
\end{center}
\caption{Dialogue for the "Extract information after the discussion" sample in Table \ref{Tab: case_analyses}}
\label{TB:}
\end{table*}

\begin{table*}[h]
\begin{center}
\resizebox{\linewidth}{!}{
\begin{tabular}{|l|l|}
\hline
\multicolumn{2}{|l|}{Phil: good evening deana ! many thanks for this nice card from you . constantine was very happy !. are these sunglasses also from you ?} \\
\multicolumn{2}{|l|}{Deana: i thought they belonged your cathreen !} \\
\multicolumn{2}{|l|}{Phil: nope . she says they aren't hers .} \\
\multicolumn{2}{|l|}{Deana: mine either . look , maybe you feel like keeping them ?. i seem to have so many sunglasses .. 8} \\
\multicolumn{2}{|l|}{Phil: where did you find them , possible that they belong to adrian ?} \\
\multicolumn{2}{|l|}{Deana: in this empty place above the radio . in the very back .. if adrian wants it , no pro !. exactly !} \\
\multicolumn{2}{|l|}{Phil: ok , they don't belong to any of us , and nobody else drove your car . but we can look after these sunglasses .} \\
\multicolumn{2}{|l|}{Deana: glad to hear it !} \\ \hline
\multicolumn{1}{|r|}{Longest-1} & \begin{tabular}[c]{@{}l@{}}Phil said good evening deana ! many thanks for this nice card from you . constantine was very happy !. \\ are these sunglasses also from you ?\end{tabular} \\ \hline
\multicolumn{1}{|r|}{BART-1} & Phil and Deana will look after Adrian's sunglasses. \\ \hline
\multicolumn{1}{|r|}{{\model-1}} & Deana found Adrian's sunglasses in the back of Phil's car. \\ \hline
\multicolumn{1}{|r|}{BART} & Phil and Deana are going to look after Adrian's sunglasses. \\ \hline
\multicolumn{1}{|r|}{\model} & Phil got a card from Deana. Deana found them in the empty place above the radio. Deana has a lot of them. \\ \hline
\multicolumn{1}{|r|}{Gold} & \begin{tabular}[c]{@{}l@{}}Phil received a card from Deana. Constantine was happy. Phil has sunglasses, that Deana found in the back above the radio.\\ Deana and Phil don't know who they belong too. Phil will keep the sunglasses.\end{tabular} \\ \hline
\end{tabular}
}
\end{center}
\caption{Test set example for qualitative study.}
\label{TB:}
\end{table*}

\begin{table*}[h]
\begin{center}
\resizebox{\linewidth}{!}{
\begin{tabular}{|r|l}
\hline
\multicolumn{2}{|l|}{Celia: where do you want to go for holiday ?} \\
\multicolumn{2}{|l|}{Mike: i was thinking about egypt} \\
\multicolumn{2}{|l|}{Celia: too hot . what about croatia ?} \\
\multicolumn{2}{|l|}{Mike: good idea , i've never been there} \\ \hline
Longest-1 & \multicolumn{1}{l|}{Celia said where do you want to go for holiday ?} \\ \hline
BART-1 & \multicolumn{1}{l|}{Mike wants to go for holiday to Egypt.} \\ \hline
{\model-1} & \multicolumn{1}{l|}{Mike wants to go on holiday to Egypt or Croatia.} \\ \hline
BART & \multicolumn{1}{l|}{Celia and Mike will go for holiday to Croatia.} \\ \hline
\model & \multicolumn{1}{l|}{Mike wants to go on holiday to Egypt. Celia thinks it's too hot. Mike has never been to Croatia, but he likes the idea.} \\ \hline
Gold & \multicolumn{1}{l|}{\begin{tabular}[c]{@{}l@{}}Mike considers going to Egypt for holiday. It's too hot for Celia, she suggests Croatia instead. Mark likes the idea, he's \\ never been there.\end{tabular}} \\ \hline
\end{tabular}
}
\end{center}
\caption{Test set example for qualitative study.}
\label{TB:}
\end{table*}

\begin{table*}[h]
\begin{center}
\resizebox{\linewidth}{!}{
\begin{tabular}{|l|l}
\hline
\multicolumn{2}{|l|}{Diane: how long do you have to work tonight ?} \\
\multicolumn{2}{|l|}{Ross: about 2 hours , why ?} \\
\multicolumn{2}{|l|}{Diane: i just wanted to do something maybe} \\
\multicolumn{2}{|l|}{Ross: i think i'll be worn out after all hat work , baby} \\
\multicolumn{2}{|l|}{Diane: we can just chill at home , don't worry. i just wanted to prepare} \\
\multicolumn{2}{|l|}{Ross: ok. then just to be safe let's say it will take me 3 hours} \\
\multicolumn{2}{|l|}{Diane: but you just said 2 !} \\
\multicolumn{2}{|l|}{Ross: ... , Diane , don't start again} \\
\multicolumn{2}{|l|}{Diane: what am i starting !. you're impossible} \\
\multicolumn{2}{|l|}{Ross: can't you understand that this is important to me !. my career depends on it !} \\
\multicolumn{2}{|l|}{Diane: well , if your career is the most important thing in the world then i wouldn't want to disturb !} \\ \hline
\multicolumn{1}{|r|}{Longest-1} & \multicolumn{1}{l|}{Diane said well , if your career is the most important thing in the world then i wouldn't want to disturb !} \\ \hline
\multicolumn{1}{|r|}{BART-1} & \multicolumn{1}{l|}{Ross has to work for 2 hours tonight.} \\ \hline
\multicolumn{1}{|r|}{{\model-1}} & \multicolumn{1}{l|}{Ross has to work 3 hours tonight.} \\ \hline
\multicolumn{1}{|r|}{BART} & \multicolumn{1}{l|}{Ross has to work tonight for 2 hours. Ross and Diane will chill at home.} \\ \hline
\multicolumn{1}{|r|}{\model} & \multicolumn{1}{l|}{Ross has to work 3 hours tonight.} \\ \hline
\multicolumn{1}{|r|}{Gold} & \multicolumn{1}{l|}{Diane is not happy with Ross prioritising work over spending time with her.} \\ \hline
\end{tabular}
}
\end{center}
\caption{Test set example for qualitative study.}
\label{TB:}
\end{table*}